\documentclass[pdflatex,sn-nature]{sn-jnl}% Style for submissions to Nature Portfolio journals
\usepackage{hyperref}
\usepackage{graphicx}%
\usepackage{multirow}%
\usepackage{amsmath,amssymb,amsfonts}%
\usepackage{amsthm}%
\usepackage{mathrsfs}%
\usepackage[title]{appendix}%
\usepackage{xcolor}%
\usepackage{textcomp}%
\usepackage{manyfoot}%
\usepackage{booktabs}%
\usepackage{algorithm}%
\usepackage{algorithmicx}%
\usepackage{algpseudocode}%
\usepackage{listings}%
\usepackage{enumitem}
\usepackage{array}         % for better column control
\usepackage{tabularx}
\usepackage{colortbl}

\theoremstyle{thmstyleone}%
\theoremstyle{thmstyletwo}%

\theoremstyle{thmstylethree}%

\begin{document}

\title[Article Title]{Consumer Attitudes Towards AI in Digital Health -- A Mixed-Methods Survey in Australia}


\author[1]{\fnm{Wei} \sur{Zhou}}
%\email{wei.zhou2@monash.edu}

\author*[1]{\fnm{Rashina} \sur{Hoda}}\email{rashina.hoda@monash.edu}

\author[2]{\fnm{Joycelyn} \sur{Ling}}
%\email{Joycelyn.Ling@dhcrc.com}

\affil*[1]{\orgdiv{Faculty of Information Technology}, \orgname{Monash University}, \orgaddress{\street{Wellington Rd}, \city{Clayton}, \postcode{3800}, \state{VIC}, \country{Australia}}}

\affil[2]{\orgname{Digital Health CRC}, \orgaddress{\street{George Street}, \city{Sydney}, \postcode{2000}, \state{NSW}, \country{Australia}}}


\abstract{
AI applications are increasingly being introduced into digital health. While technical performance has advanced rapidly, successful deployment mainly depends on consumer attitudes, especially to patient-facing applications. However, most existing research examines consumer attitudes towards healthcare AI at an abstract level rather than in response to concrete artefacts.
We report a mixed-methods survey study in Australia (N=275) examining consumer readiness, acceptance, trust, and risk perceptions of healthcare AI, combined with a scenario-based evaluation of an AI-generated versus clinician-written consultation summary. 
Participants expressed moderate optimism and strong perceived usefulness and ease of use, but also substantial concerns about accuracy, safety, and data use. In the scenario task, the AI-generated summary was strongly preferred for quality, empathy, and overall usefulness, yet identification of the AI summary was near chance. 
Findings show that consumers judge AI through concrete communication quality and visible human governance, underscoring the need for clinically supervised deployment frameworks beyond technical performance alone.
}


\maketitle

\section{Introduction}\label{introduction}

Artificial intelligence (AI), and particularly, large language models (LLMs), are rapidly being integrated into digital health to support clinical documentation, decision support, patient communication, and administrative workflows \cite{bajwa2021artificial,al2023review,rashid2025paradigm,kwong2024integrating,silcox2024potential}. Among these applications, automated clinical summarization has emerged as one of the most immediately deployable and operationally attractive use cases, promising to reduce documentation burden while improving consumers\footnote{In this paper, we use the term \emph{consumers} to refer to mainly patients and carers, and health service users; Healthcare professionals or clinicians are not considered consumers.} access to and understanding of their own health information \cite{chen2024exploring,lee2024prospects,clough2024transforming}. Recent studies have demonstrated substantial technical progress in generating discharge summaries, consultation notes, and structured clinical narratives using generative AI systems, and pilot deployments are already underway in real-world clinical environments \cite{chua2024integration,shemtob2025comparing,fraile2025expert,koh2025using}.

However, successful integration of AI into digital health depends not only on technical performance, but also on social legitimacy, trust, and acceptance by consumers \cite{young2021patient,esmaeilzadeh2021patients,nuccetelli2025use}. Healthcare is a high-stakes, trust-sensitive domain in which errors, opacity, or misalignment with consumer values can have serious consequences \cite{williamson2024balancing,who_ai_health}. A growing body of research shows that consumer attitudes towards healthcare AI are neither uniformly optimistic nor uniformly resistant, but rathercharacterized by a complex mixture of perceived benefits, concerns about safety and privacy, expectations of human oversight, and fears of dehumanization of care \cite{gundlack2025patients,foresman2025patient,reis2025public,ding2025trust}.

Recent work further suggests that attitudes towards AI in healthcare are highly context-dependent \cite{cinalioglu2023exploring,reis2024influence,changes_chatgpt_perception}. People evaluate AI differently depending on whether it is used for diagnosis, triage, decision support, documentation, or patient-facing communication \cite{kwong2024integrating,silcox2024potential}. Even when AI is positioned as an assistive tool rather than a replacement for clinicians, its presence can influence how consumers perceive clinical competence, responsibility, and trustworthiness \cite{reis2025public,ding2025trust} of AI-supported care. Moreover, while AI-generated text can sometimes be perceived as empathetic or supportive, such judgements are highly sensitive to tone, framing, and narrative style \cite{chen2025patient,ovsyannikova2025third}.

Despite this growing literature, most existing studies examine attitudes towards AI in healthcare at a relatively abstract level or focus primarily on clinician-facing systems such as diagnostic support or workflow automation \cite{young2021patient,nuccetelli2025use}. Comparatively little empirical work has examined how consumers respond to concrete, realistic AI-generated artefacts that they might directly receive in routine care, such as consultation summaries \cite{chen2024exploring,clough2024transforming,shemtob2025comparing}. In particular, there is limited evidence on how consumers evaluate such summaries simultaneously across multiple dimensions, including clinical relevance, clarity, empathy, usability, trustworthiness, and perceived authorship \cite{goodman2024ai,fraile2025expert}.

At the same time, policy and regulatory frameworks for AI in healthcare are rapidly evolving, including in Australia and comparable jurisdictions \cite{aus_ai_ethics,aus_ai_assurance,tga_ai_md,who_ai_health,oecd_ai}. These frameworks emphasize safety, clinical validation, accountability, and data governance \cite{lekadir2025future}. While such principles are essential, there remains limited empirical evidence on how consumers evaluate patient-facing generative AI outputs in practice, particularly for systems that mediate understanding, reassurance, and sense-making rather than merely supporting technical clinical decisions \cite{gundlack2025patients,foresman2025patient}.

To address this gap, this study reports a mixed-methods investigation of consumer attitudes towards AI in digital health, with a specific focus on AI-generated consultation summaries as a realistic and increasingly likely patient-facing use case. We combine (i) established quantitative measures of technology readiness, acceptance, trust, and risk perception, (ii) a scenario-based evaluation in which participants compare an AI-generated and a clinician-written consultation summary across dimensions of quality, empathy, preference, and perceived authorship, and (iii) qualitative analysis of participants’ explanations, expectations, and concerns. By grounding the study in concrete artefacts rather than abstract descriptions of AI, and by foregrounding consumer experience rather than purely technical performance, this work aims to generate actionable evidence to inform the responsible design, governance, and evaluation of patient-facing generative AI systems in healthcare.
Although situated in Australia, these findings contribute to broader international efforts to align patient-facing AI with consumer expectations and trustworthy governance.

\begin{table}[t]
\centering
\caption{Demographic characteristics of survey participants (N = 275).}
\label{tab:demographics}
\begin{tabular}{llrr}
\toprule
\textbf{Characteristics} & \textbf{Group} & \textbf{Number of participants} & \textbf{Percentage} \\
\midrule

\multirow{6}{*}{Age group} 
 & 18--24 & 53 & 19.3\% \\
 & 25--34 & 57 & 20.7\% \\
 & 35--44 & 74 & 26.9\% \\
 & 45--54 & 40 & 14.5\% \\
 & 55--64 & 33 & 12.0\% \\
 & 65+ & 18 & 6.5\% \\

\midrule
\multirow{4}{*}{Gender}
 & Man & 138 & 50.2\% \\
 & Woman & 133 & 48.4\% \\
 & Non-binary & 3 & 1.1\% \\
 & Prefer not to say & 1 & 0.4\% \\

\midrule
\multirow{8}{*}{State / Territory}
 & VIC & 129 & 46.9\% \\
 & NSW & 62 & 22.5\% \\
 & QLD & 38 & 13.8\% \\
 & WA & 17 & 6.2\% \\
 & SA & 14 & 5.1\% \\
 & ACT & 9 & 3.3\% \\
 & TAS & 5 & 1.8\% \\
 & Prefer not to say & 1 & 0.4\% \\

\midrule
\multirow{4}{*}{Area type}
 & Metropolitan & 205 & 74.5\% \\
 & Regional & 50 & 18.2\% \\
 & Rural & 17 & 6.2\% \\
 & Prefer not to say & 3 & 1.1\% \\

\midrule
\multirow{6}{*}{Highest Education}
 & High school & 41 & 14.9\% \\
 & Certificate/Diploma & 37 & 13.5\% \\
 & Bachelor & 91 & 33.1\% \\
 & Master & 59 & 21.5\% \\
 & PhD or equivalent & 45 & 16.4\% \\
 & Other & 2 & 0.7\% \\

\bottomrule
\end{tabular}
\end{table}

\begin{table}[t]
\centering
\caption{Participants' technology background, experience with non-health AI, and trust on AI in healthcare (N = 275).}
\label{tab:attitudes}
\begin{tabular}{llrr}
\toprule
\textbf{Category} & \textbf{Response} & \textbf{N} & \textbf{\%} \\
\midrule

\multirow{3}{*}{Technology skill}
 & Advanced & 135 & 49.1\% \\
 & Intermediate & 124 & 45.1\% \\
 & Basic & 16 & 5.8\% \\

\midrule

\multirow{4}{*}{Attitude to technology}
 & I enjoy trying new technology & 155 & 56.4\% \\
 & I feel neutral about it & 80 & 29.1\% \\
 & I use it only when necessary & 38 & 13.8\% \\
 & I avoid it whenever possible & 2 & 0.7\% \\

\midrule
\multirow{7}{*}{Non-health AI usage frequency}
 & Several times a day & 101 & 36.7\% \\
 & About once a day & 34 & 12.4\% \\
 & A few times a week & 56 & 20.4\% \\
 & About once a week & 25 & 9.1\% \\
 & A few times a month & 22 & 8.0\% \\
 & Rarely ($<$ once a month) & 27 & 9.8\% \\
 & Never & 10 & 3.6\% \\

% \midrule
% \multirow{8}{*}{Experience with AI in healthcare$^\dagger$}
%  & AI-based health monitoring devices or wearables & 116 & 42.2\% \\
%  & AI-based virtual health assistants & 109 & 39.6\% \\
%  & AI-based health management tools & 59 & 21.5\% \\
%  & AI-based transcription of medical consultations & 48 & 17.5\% \\
%  & AI-based mental health support apps & 38 & 13.8\% \\
%  & AI-based diagnostic imaging or assessment & 14 & 5.1\% \\
%  & Not sure / have not knowingly used AI & 77 & 28.0\% \\
%  & Other & 2 & 0.7\% \\
\midrule

\multirow{5}{*}{Current trust level of AI in healthcare}
 & Very high & 6 & 2.2\% \\
 & High & 39 & 14.2\% \\
 & Moderate & 142 & 51.6\% \\
 & Low & 66 & 24.0\% \\
 & Very low & 22 & 8.0\% \\
\bottomrule
\end{tabular}

\vspace{1mm}
% \footnotesize $^\dagger$Multiple responses were allowed; percentages do not sum to 100\%.
\end{table}

\section{Results}\label{result}

\subsection{Background information of the participants}

A total of $N = 275$ responses were included in the final analysis after standard data quality checks (e.g., removal of incomplete responses or quickly completed responses, and failed attention checks). 

Table~\ref{tab:demographics} summarises the demographic characteristics of the 275 survey participants. 
\begin{itemize}
    \item The sample covered a broad age range, with the largest proportion of participants aged 35--44 years (26.9\%), followed by those aged 25--34 years (20.7\%) and 18--24 years (19.3\%).
    Compared to Australian age distributions in the 2021 Census\footnote{\url{https://www.abs.gov.au/statistics/people/people-and-communities/snapshot-australia/latest-release}}, younger and middle-aged adults (18-44 years) were more prominent in our sample than older adults, a pattern commonly observed in the recruitment of online participants.
    \item Gender distribution was balanced, with 50.2\% identifying as men and 48.4\% as women.
    \item Most participants resided in metropolitan areas (74.5\%), and the largest proportions were from Victoria (46.9\%) and New South Wales (22.5\%).
    \item The sample was relatively highly educated, with more than two-thirds holding at least a bachelor’s degree (71.0\%), including 37.9\% with a postgraduate qualification.
\end{itemize}

Table~\ref{tab:attitudes} reports participants’ technology background, experience with non-health AI, and current level of trust in AI in healthcare.

Nearly all participants rated their technology skills as intermediate or advanced (94.2\%), and the majority reported enjoying trying new technologies (56.4\%). Most participants reported frequent use of non-health AI applications, with 49.1\% indicating usage at least once per day.

%In terms of experience with AI in healthcare, the most commonly reported exposures were AI-based health monitoring devices or wearables (42.2\%) and virtual health assistants (39.6\%), although 28.0\% of participants indicated that they were not sure or had not knowingly used AI in healthcare. 

Regarding general trust, over half of the participants reported a moderate level of trust in AI in healthcare (51.6\%), while 32.0\% reported low or very low trust and only 16.4\% reported high or very high trust.
In this quantitative result, participants showed a distribution of moderate to low levels of trust with substantial variance. The qualitative findings explain this dispersion by revealing that trust is not a single attitude, but is composed of several distinct concerns.
\begin{itemize}
    \item Participants explicitly pointed to the risk of clinical errors (``\textit{AI could make mistakes, which makes it unreliable for health care}''), the immaturity of the technology (``\textit{still in its infancy so can't say I fully trust it}''), and broader uncertainty about reliability.
    \item At the same time, some expressed conditional trust mediated by care provision context and clinicians (``\textit{I definitely don't have high trust in AI in healthcare, but I trust healthcare providers to choose AI technology that is ethical}''). This helps explain why the quantitative trust scores cluster around the middle of the scale rather than at the extremes: participants are not rejecting AI outright, but are evaluating it under conditions of uncertainty and risk.
    \item Crucially, many participants framed acceptability in terms of human oversight, stating for example ``\textit{If the doctor reviews them before they are available to the patient}'', ``\textit{Healthcare provider should review all the AI-generated summary}'', and ``\textit{Having the information validated by a human}''. This supports the interpretation of the quantitative trust results as reflecting conditional and context-dependent acceptance rather than simple approval or disapproval.
\end{itemize}

These findings are broadly consistent with international consumer surveys, which similarly report that digitally engaged participants often have prior exposure to everyday AI applications and express moderate, conditional trust in healthcare AI \cite{young2021patient,chew2022perceptions,cinalioglu2023exploring}.

\begin{figure*}[t]
    \centering
    \includegraphics[width=\textwidth]{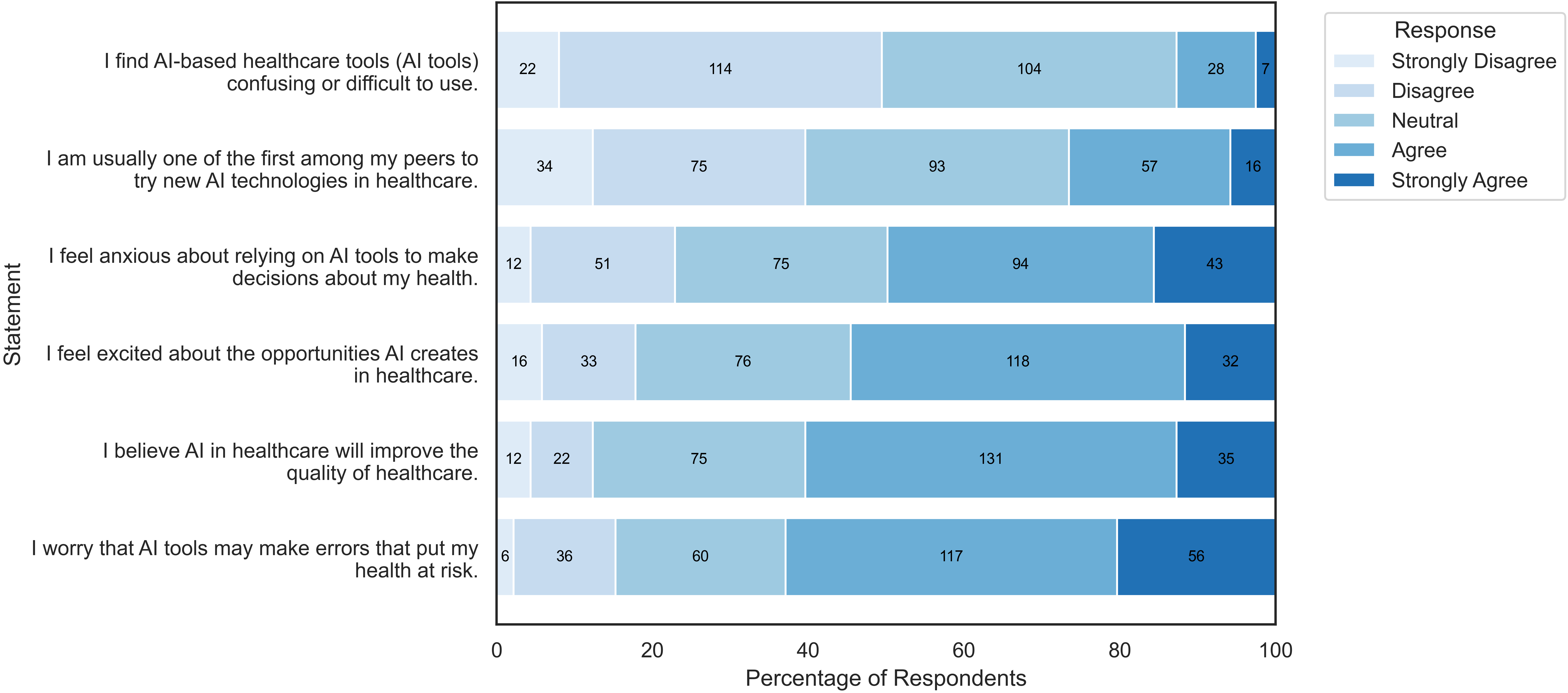}
    \caption{Statistics of TRI Statements}
    \label{fig_att1}
\end{figure*}

\subsection{Technology Readiness Index (TRI) of AI in Healthcare}

Figure~\ref{fig_att1} provides a descriptive analysis of participant technology readiness regarding AI in healthcare, based on the core dimensions of the Technology Readiness Index: Optimism and Innovativeness (Positive Drivers), and Discomfort and Insecurity (Negative Inhibitors). Mean scores are calculated on a scale from Strongly Disagree ($-2$) to Strongly Agree ($+2$), where a positive value indicates overall agreement with the statement. The analysis is summarized in Table~\ref{tab:tri_group1}.

\begin{table}[t]
    \centering
    \caption{Analysis of TRI Statements (N=275)}
    \label{tab:tri_group1}
    \begin{tabular}{p{4.5cm} c c c c}
        \toprule
        \textbf{Statement} & \textbf{Mean Score (M)} & \textbf{Agree/SA} ($\%$) & \textbf{Disagree/SD} ($\%$) & \textbf{Neutral} ($\%$) \\
        \midrule
        \multicolumn{5}{l}{\textit{\textbf{Positive Drivers}}} \\
        \midrule
        I believe AI in healthcare will improve the quality of healthcare. & $0.56$ & $60.4$ & $12.4$ & $27.3$ \\
        \\
        I feel excited about the opportunities AI creates in healthcare. & $0.43$ & $54.5$ & $17.8$ & $27.6$ \\
        \\
        I am usually one of the first among my peers to try new AI technologies in healthcare. & $-0.20$ & $26.5$ & $39.6$ & $33.8$ \\
        \midrule
        \multicolumn{5}{l}{\textit{\textbf{Negative Inhibitors}}} \\
        \midrule
        I find AI-based healthcare tools (AI tools) confusing or difficult to use. & $-0.42$ & $12.7$ & $49.5$ & $37.8$ \\
        \\
        I feel anxious about relying on AI tools to make decisions about my health. & $0.38$ & $49.8$ & $22.9$ & $27.3$ \\
        \\
        I worry that AI tools may make errors that put my health at risk. & $0.66$ & $62.9$ & $15.3$ & $21.8$ \\
        \bottomrule
    \end{tabular}
    \footnotesize  \quad SA denotes Strongly Agree; SD denotes Strongly Disagree.
\end{table}

\subsubsection{Analysis of Positive Drivers}

\textbf{Optimism} towards AI in healthcare was \textbf{moderate}. The statement, ``\textit{I believe AI in healthcare will improve the quality of healthcare}'',  recorded a positive mean score ($\mathbf{M = 0.56}$), with $\mathbf{60.4\%}$ of participants expressing agreement (Agree or Strongly Agree). Participants also reported positive affect towards AI, with $\mathbf{54.5\%}$ agreeing that they ``\textit{feel excited about the opportunities AI creates in healthcare}'' ($\mathbf{M = 0.43}$). However, self-reported early adoption was limited: the statement, ``\textit{I am usually one of the first among my peers to try new AI technologies in healthcare}'', returned a slightly negative mean score ($\mathbf{M = -0.20}$), with a larger proportion disagreeing ($\mathbf{39.6\%}$) than agreeing ($\mathbf{26.5\%}$). This suggests that while consumers are generally optimistic and interested, many do not view themselves as early adopters.

\subsubsection{Analysis of Negative Inhibitors}

Participants generally reported  \textbf{low discomfort} with usability: the statement ``\textit{I find AI-based healthcare tools confusing or difficult to use}'' had a negative mean score ($\mathbf{M = -0.42}$), with nearly half disagreeing ($\mathbf{49.5\%}$) and only $\mathbf{12.7\%}$ agreeing, indicating that most do \emph{not} anticipate usability confusion.

In contrast, \textbf{insecurity} related to safety and reliance was more pronounced. Nearly half agreed that \textit{they would feel anxious relying on AI tools to make health decisions} ($\mathbf{49.8\%}$ agreement; $\mathbf{M = 0.38}$). Concern about AI errors was even stronger: $\mathbf{62.9\%}$ agreed that \textit{they worry AI tools may make errors that put their health at risk} ($\mathbf{M = 0.66}$). Together, these results indicate that the primary inhibitors are not so much usability barriers, but perceived risk and the consequences of potential AI errors.

Overall, participants demonstrate a mixed technology readiness profile: moderate optimism and excitement co-exist with a cautious stance towards being early adopters of new AI technologies. This pattern suggests that adoption efforts should emphasize robust clinical governance, clear safety assurances, and human oversight \cite{lekadir2025future, jindal2024ensuring}, while also building consumer confidence through transparent communication and exposure to well-integrated, easy-to-use AI-enabled workflows \cite{reis2024influence, ding2025trust, kwong2024integrating}.

\subsection{Technology Acceptance Model (TAM) of AI Summarization Tools}

\begin{figure*}[t]
    \centering
    \includegraphics[width=\textwidth]{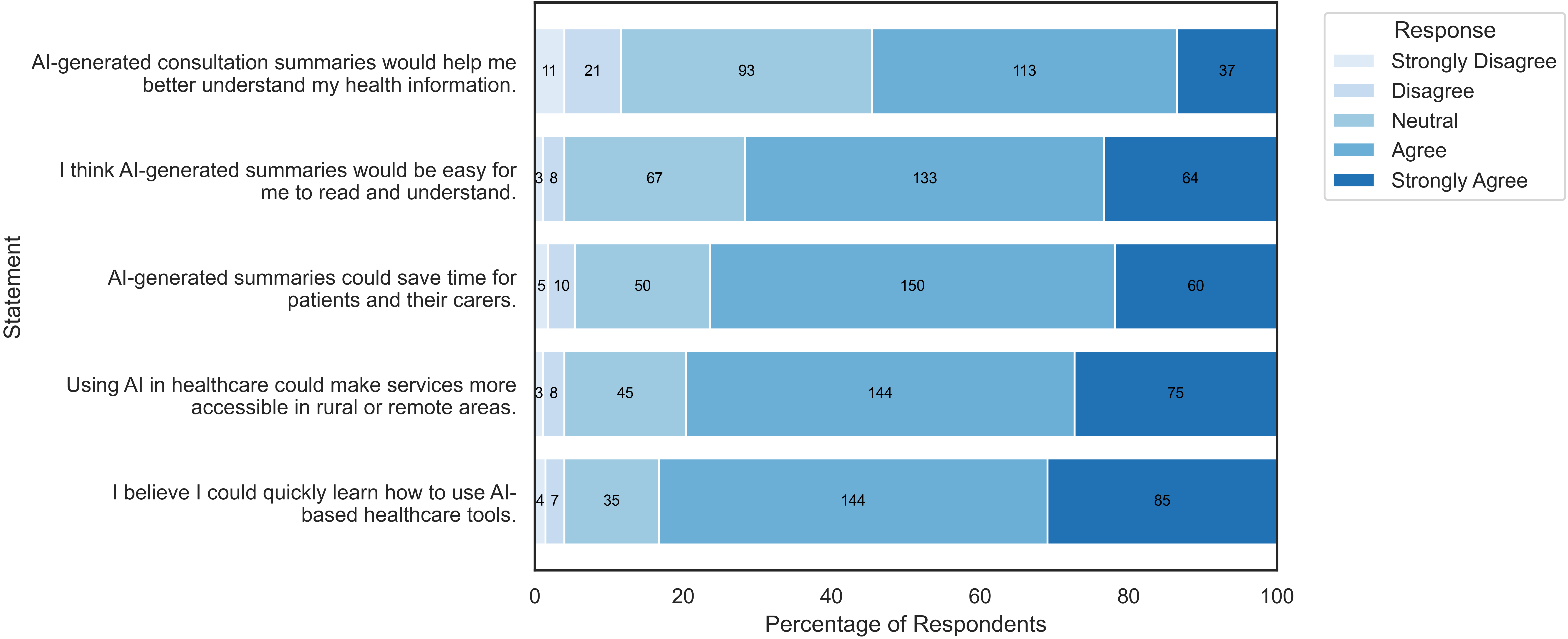}
    \caption{Statistics of TAM Statements}
    \label{fig_att2}
\end{figure*}

Figure~\ref{fig_att2} provides a detailed breakdown of consumer perceptions regarding the utility and usability of AI in healthcare, based on the established constructs of the Technology Acceptance Model. The analysis is summarized in Table~\ref{tab:tam_group2}.

\begin{table}[h!]
    \centering
    \caption{Analysis of TAM Statements (N=275)}
    \label{tab:tam_group2}
    \begin{tabular}{p{4.5cm} c c c c}
        \toprule
        \textbf{Statement} & \textbf{Mean Score} & \textbf{Agree/SA} ($\%$) & \textbf{Disagree/SD} ($\%$) & \textbf{Neutral} ($\%$) \\
        \midrule
        \multicolumn{5}{l}{\textit{\textbf{Perceived Usefulness (PU)}}} \\
        \midrule
        AI-generated summaries could save time for patients and their carers. & $0.91$ & $76.4$ & $5.5$ & $18.2$ \\
        \\
        AI-generated consultation summaries would help me better understand my health information. & $0.52$ & $54.5$ & $11.6$ & $33.8$ \\
        \\
        Using AI in healthcare could make services more accessible in rural or remote areas. & $1.02$ & $79.6$ & $4.0$ & $16.4$ \\
        \midrule
        \multicolumn{5}{l}{\textit{\textbf{Perceived Ease of Use (PEOU)}}} \\
        \midrule
        I think AI-generated summaries would be easy for me to read and understand. & $0.90$ & $71.6$ & $4.0$ & $24.4$ \\
        \\
        I believe I could quickly learn how to use AI-based healthcare tools. & $1.09$ & $83.3$ & $4.0$ & $12.7$ \\
        \bottomrule
    \end{tabular}
\end{table}

\subsubsection{Analysis of Perceived Usefulness (PU)}

\textbf{Perceived Usefulness} emerged as a strong driver of acceptance for AI in healthcare, particularly for accessibility and efficiency outcomes. The highest utility endorsement related to improving access: $\mathbf{79.6\%}$ agreed that \textit{AI could make services more accessible in rural or remote areas }($\mathbf{M = 1.02}$). Time-saving potential was also strongly supported ($\mathbf{76.4\%}$; $\mathbf{M = 0.91}$), indicating a clear expectation of workflow and information-delivery benefits for consumers.

In contrast, perceived usefulness for improving \textbf{understanding} of health information was positive but more uncertain. While a majority still agreed ($\mathbf{54.5\%}$; $\mathbf{M = 0.52}$), one-third of participants selected Neutral ($\mathbf{33.8\%}$), suggesting that consumers may be less certain about the extent to which AI summaries will clarify complex clinical content.

\subsubsection{Analysis of Perceived Ease of Use (PEOU)}

\textbf{Perceived Ease of Use} was very strong. Most participants believed \textit{AI-generated summaries would be easy to read and understand} ($\mathbf{71.6\%}$ agreement; $\mathbf{M = 0.90}$). Confidence in learning to use AI-based healthcare tools was even higher ($\mathbf{83.3\%}$ agreement; $\mathbf{M = 1.09}$), indicating that usability and learnability are unlikely to be major barriers to adoption.

In summary, consumers show strong acceptance of both perceived usefulness (especially for accessibility and time-saving) and perceived ease of use (readability and learnability). The primary nuance is that perceived usefulness for improving personal understanding of health information is positive but less decisive, indicating an opportunity to improve trust and clarity through careful design, plain-language summaries, and clinician-supported explanations \cite{goodman2024ai, borges2023barriers,lee2024prospects, chen2024exploring}.

\subsection{Consumer Attitudes and Behavioral Intention towards AI Summarization}

\begin{figure*}[t]
    \centering
    \includegraphics[width=\textwidth]{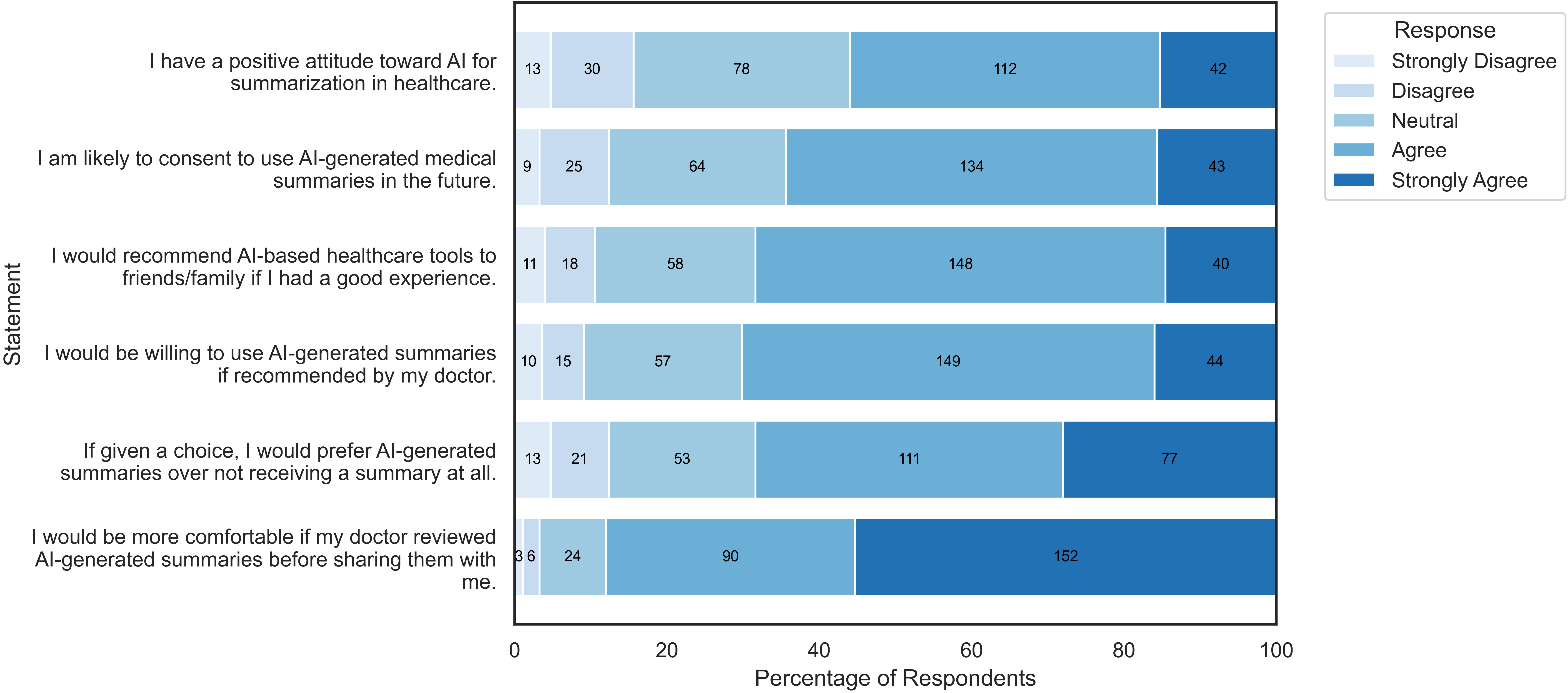}
    \caption{Statements on Attitudes and Behavioral Intention towards AI Summarization}
    \label{fig_att3}
\end{figure*}

Figure~\ref{fig_att3} examines consumer attitudes and indicators of behavioral intention towards the use of AI for summarization in healthcare. The analysis explores general disposition, willingness to use, future consent likelihood, and the influence of clinician endorsement. The results are summarized in Table~\ref{tab:attitudes_group3}.

\subsubsection{Analysis of Attitudes and Acceptance}

Consumers demonstrated a clear, although conditional, positive orientation towards AI summarization.

The statement reflecting the foundational \textbf{Attitude} (``\textit{I have a positive attitude towards AI for summarization in healthcare}'') yielded a positive mean score ($\mathbf{M = 0.51}$), with $\mathbf{56.0\%}$ agreement. Notably, $\mathbf{28.4\%}$ of participants remained neutral, indicating that a significant segment holds a non-committal or still-forming attitude.

However, once key conditional factors are introduced, the \textbf{Behavioural Intention} strengthens considerably:
\begin{itemize}
    \item \textbf{Preference and Consent:} $\mathbf{68.4\%}$ of participants indicated they would ``\textit{prefer AI-generated summaries over not receiving a summary at all}'' ($\mathbf{M = 0.79}$), establishing AI summaries as a desirable alternative to the current default. Similarly, the likelihood to consent to use AI summaries in the future was strong ($\mathbf{M = 0.64}$, $\mathbf{64.4\%}$ agreement), suggesting a high propensity for future adoption.
    \item \textbf{Social Influence:} The willingness to use AI summaries if ``\textit{recommended by my doctor}'' achieved a high mean score of $\mathbf{M = 0.73}$ and $\mathbf{70.2\%}$ agreement. Following this, the intention to become an advocate, or ``\textit{recommend AI-based healthcare tools to friends/family},'' was also robust ($\mathbf{M = 0.68}$, $\mathbf{68.4\%}$ agreement). These results highlight the critical role of clinician endorsement and positive user experience in driving adoption.
\end{itemize}

\begin{table}[t]
    \centering
    \caption{Analysis of Statements on Attitudes and Behavioural Intention towards AI Summarization (N=275)}
    \label{tab:attitudes_group3}
    \begin{tabular}{p{4.5cm} c c c c}
        \toprule
        \textbf{Statement} & \textbf{Mean Score} & \textbf{Agree/SA} ($\%$) & \textbf{Disagree/SD} ($\%$) & \textbf{Neutral} ($\%$) \\
        \midrule
        I would be more comfortable if my doctor reviewed AI-generated summaries before sharing them with me. & $1.39$ & $88.0$ & $3.3$ & $8.7$ \\
        \\
        If given a choice, I would prefer AI-generated summaries over not receiving a summary at all. & $0.79$ & $68.4$ & $12.4$ & $19.3$ \\
        \\
        I would be willing to use AI-generated summaries if recommended by my doctor. & $0.73$ & $70.2$ & $9.1$ & $20.7$ \\
        \\
        I would recommend AI-based healthcare tools to friends/family if I had a good experience. & $0.68$ & $68.4$ & $10.5$ & $21.1$ \\
        \\
        I am likely to consent to use AI-generated medical summaries in the future. & $0.64$ & $64.4$ & $12.4$ & $23.3$ \\
        \\
        I have a positive attitude towards AI for summarization in healthcare. & $0.51$ & $56.0$ & $15.6$ & $28.4$ \\
        \bottomrule
    \end{tabular}
\end{table}

\subsubsection{The Crucial Role of Clinical Review}

The strongest consensus across the survey was observed in the conditionality of comfort and trust. The statement, ``\textit{I would be more comfortable if my doctor reviewed AI-generated summaries before sharing them with me}'', obtained a near-unanimous mean score ($\mathbf{M = 1.39}$), with $\mathbf{88.0\%}$ of participants in agreement or strong agreement. This finding represents a key prerequisite for adoption, demonstrating that while consumers recognize utility, human oversight -- specifically, clinician review -- is essential for maximizing comfort.

The data affirms that consumers hold a favorable disposition towards AI summarization, which translates into high \textbf{Behavioural Intention} across key indicators (use, consent, recommendation). This intention is maximized when the technology is presented as a valuable option (preferable to no summary) and is strongly mediated by professional endorsement (clinician recommendation). Fundamentally, the path to high adoption requires clinician oversight and validation, establishing a hybrid human -- AI model of care that addresses consumer trust and comfort.

\begin{figure*}[t]
    \centering
    \includegraphics[width=\textwidth]{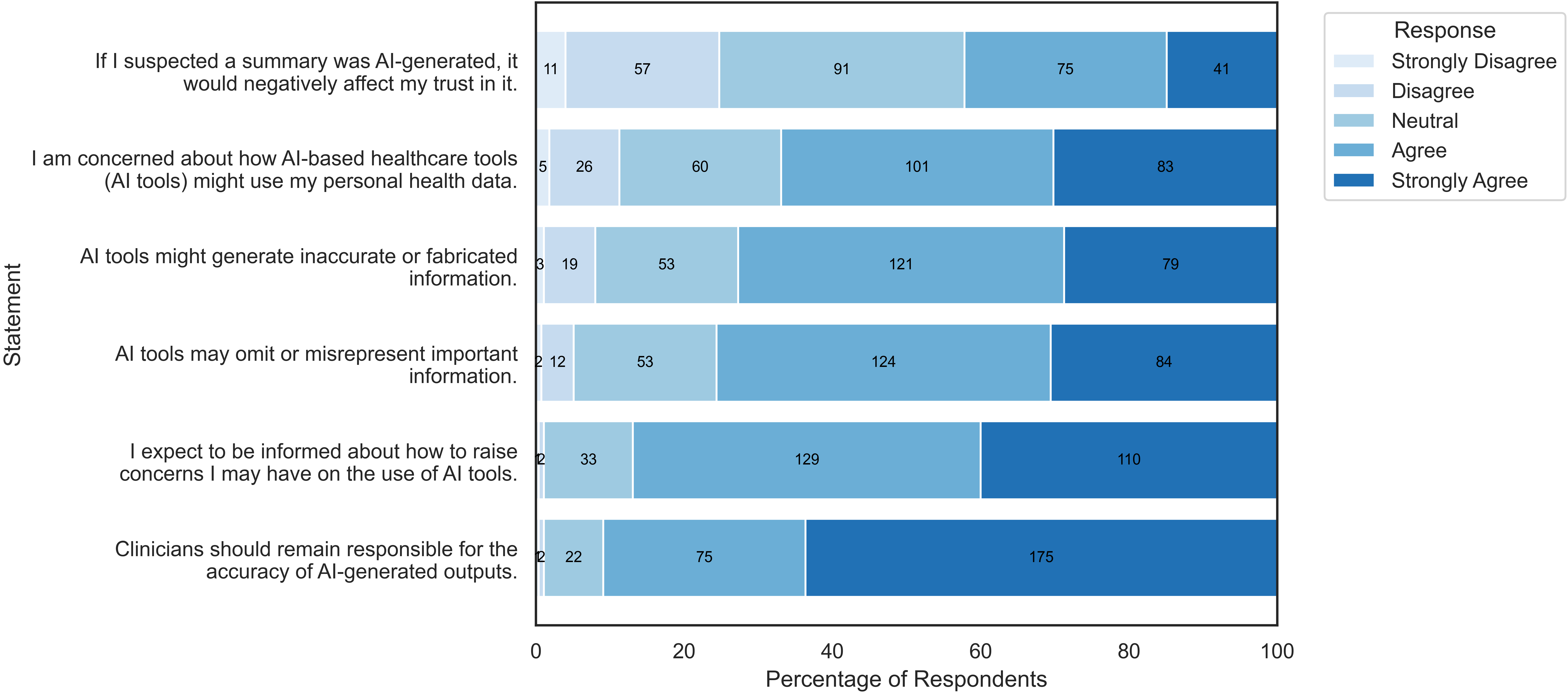}
    \caption{Statements on Risk, Trust, and Ethical Requirements}
    \label{fig_att4}
\end{figure*}

\subsection{Consumer Perceptions of Risk, Trust, and Ethical Requirements for AI in Healthcare}

Figure~\ref{fig_att4} depicts consumer concerns related to the safety, trustworthiness, and ethical governance of AI-based healthcare tools. Statements address perceived risks, the conditional nature of trust, and expectations regarding professional responsibility and transparency. The results are summarized in Table~\ref{tab:risk_trust_group4}.

\begin{table}[h!]
    \centering
    \caption{Analysis of Statements on Risk, Trust, and Ethical Concerns (N=275)}
    \label{tab:risk_trust_group4}
    \begin{tabular}{p{4.5cm} c c c c}
        \toprule
        \textbf{Statement} & \textbf{Mean Score} & \textbf{Agree/SA} ($\%$) & \textbf{Disagree/SD} ($\%$) & \textbf{Neutral} ($\%$) \\
        \midrule
        Clinicians should remain responsible for the accuracy of AI-generated outputs. & $1.53$ & $90.9$ & $1.1$ & $8.0$ \\
        \\
        I expect to be informed about how to raise concerns I may have on the use of AI tools. & $1.25$ & $86.9$ & $1.1$ & $12.0$ \\
        \\
        AI tools may omit or misrepresent important information. & $1.00$ & $75.6$ & $5.1$ & $19.3$ \\
        \\
        AI tools might generate inaccurate or fabricated information. & $0.92$ & $72.7$ & $8.0$ & $19.3$ \\
        \\
        I am concerned about how AI-based healthcare tools (AI tools) might use my personal health data. & $0.84$ & $66.9$ & $11.3$ & $21.8$ \\
        \\
        If I suspected a summary was AI-generated, it would negatively affect my trust in it. & $0.28$ & $42.2$ & $24.7$ & $33.1$ \\
        \bottomrule
    \end{tabular}
\end{table}

\subsubsection{Analysis of Perceived Risks}

Perceived risks related to AI accuracy were very high, with strong agreement across technical and information-integrity concerns.
\begin{itemize}
    \item \textbf{Accuracy Risks:} The strongest risk-related endorsement concerned omission or misrepresentation of important information ($\mathbf{M = 1.00}$), with $\mathbf{75.6\%}$ agreement. Concern that \textit{AI tools might generate inaccurate or fabricated information} was similarly high ($\mathbf{M = 0.92}$; $\mathbf{72.7\%}$ agreement), indicating substantial skepticism regarding reliability and completeness of AI outputs.
    \item \textbf{Data Privacy Risk:} Concerns over personal health data use were also prominent, with $\mathbf{66.9\%}$ agreement ($\mathbf{M = 0.84}$).
\end{itemize}

\subsubsection{Analysis of Ethical and Trust Expectations}

Consumers reported very high expectations regarding accountability and transparency.
\begin{itemize}
    \item \textbf{Accountability:} The highest mean score observed across the survey was for clinician responsibility for AI output accuracy ($\mathbf{M = 1.53}$), with $\mathbf{90.9\%}$ agreement, establishing clinical accountability as an essential requirement for AI implementation.
    \item \textbf{Transparency and Recourse:} A large majority expected to be informed about how to raise concerns regarding AI use ($\mathbf{M = 1.25}$; $\mathbf{86.9\%}$ agreement), signaling a demand for governance and clear redress mechanisms.
\end{itemize}

\subsubsection{Trust Conditionality}

The statement ``\textit{If I suspected a summary was AI-generated, it would negatively affect my trust in it}'' produced a more nuanced pattern ($\mathbf{M = 0.28}$). While $\mathbf{42.2\%}$ agreed, $\mathbf{33.1\%}$ remained neutral and $\mathbf{24.7\%}$ disagreed. This suggests that AI generation alone does not universally undermine trust, but instead introduces \emph{conditionality} depending on context and safeguards \cite{young2021patient,ding2025trust,reis2024influence}..

The findings demonstrate a clear tension: consumers perceive meaningful benefits and usability, yet simultaneously express substantial concerns about accuracy and data privacy. The exceptionally high consensus on \textbf{clinician responsibility} ($\mathbf{M = 1.53}$) and the expectation of the availability of clear pathways for explanation, correction, or escalation when AI outputs are perceived as inaccurate or harmful ($\mathbf{M = 0.84}$), indicates that consumers seek to mitigate these perceived risks through strong clinical oversight and transparent governance. Successful deployment of AI in healthcare therefore hinges on robust human oversight and clear accountability frameworks.

\begin{figure*}[t]
    \centering
    \includegraphics[width=1.2\textwidth]{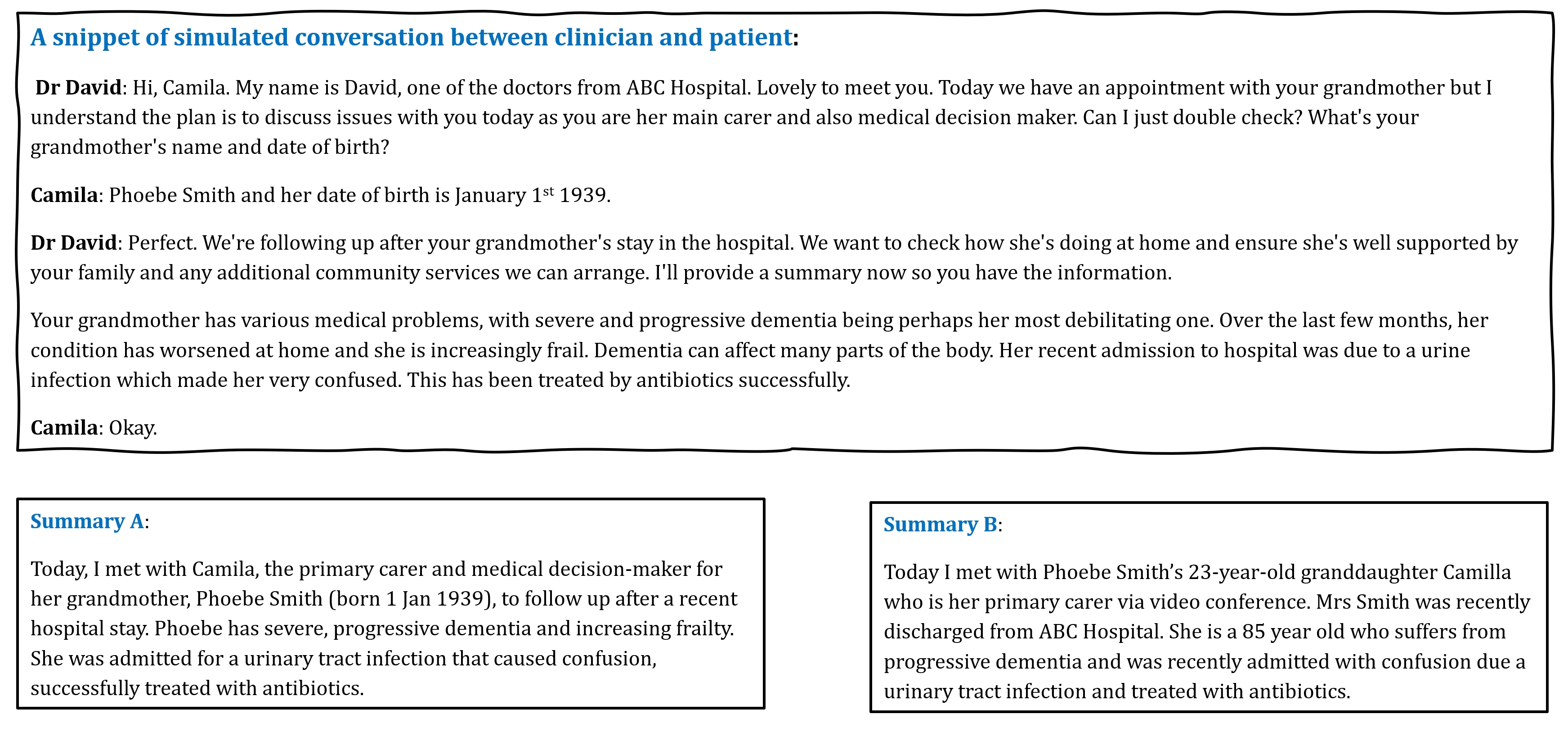}
    \caption{Scenario based consultation summarization: Summary A was generated by AI; Summary B was generated by a human clinician. Authorship information was not disclosed to participants so as not to bias them.
}
    \label{fig_scenario}
\end{figure*}

\subsection{Scenario-based comparison of AI Summary and Human Summary}

Participants also evaluated two alternative consultation summaries (Summary A generated by AI, and Summary B generated by a human clinician, as shown in Figure~\ref{fig_scenario}) across four dimensions: \textit{which summary better captured the key issues discussed during the consultation}, \textit{which felt more empathetic or patient-centred}, \textit{which they would prefer to receive as a patient or carer}, and \textit{which they believed was AI-generated}.

The clinician-written summary was prepared by a practising general practitioner with clinical experience in producing patient-facing consultation documentation. The clinician was provided with a transcript of the consultation conversation (``CONV'') and asked to write a routine post-visit summary intended for patient reference.
The AI-generated summary was produced using GPT-4 (OpenAI) with a fixed prompt designed to emulate a clinician documentation task: ``\textit{You are a professional GP doctor, and have just finished your consultation. Now you need to write up a consultation summary for the patient so that they can refer back to it when they get home. Please summarise the following conversation into a consultation report}: + CONV''.
To ensure feasibility within the survey format, we used a short excerpt of the  consultation conversation and the corresponding paired summaries. This scenario-based design was intended to support controlled comparison, although it does not capture the full variability of real-world clinical encounters.

\begin{figure*}[t]
    \centering
    \includegraphics[width=1.1\textwidth]{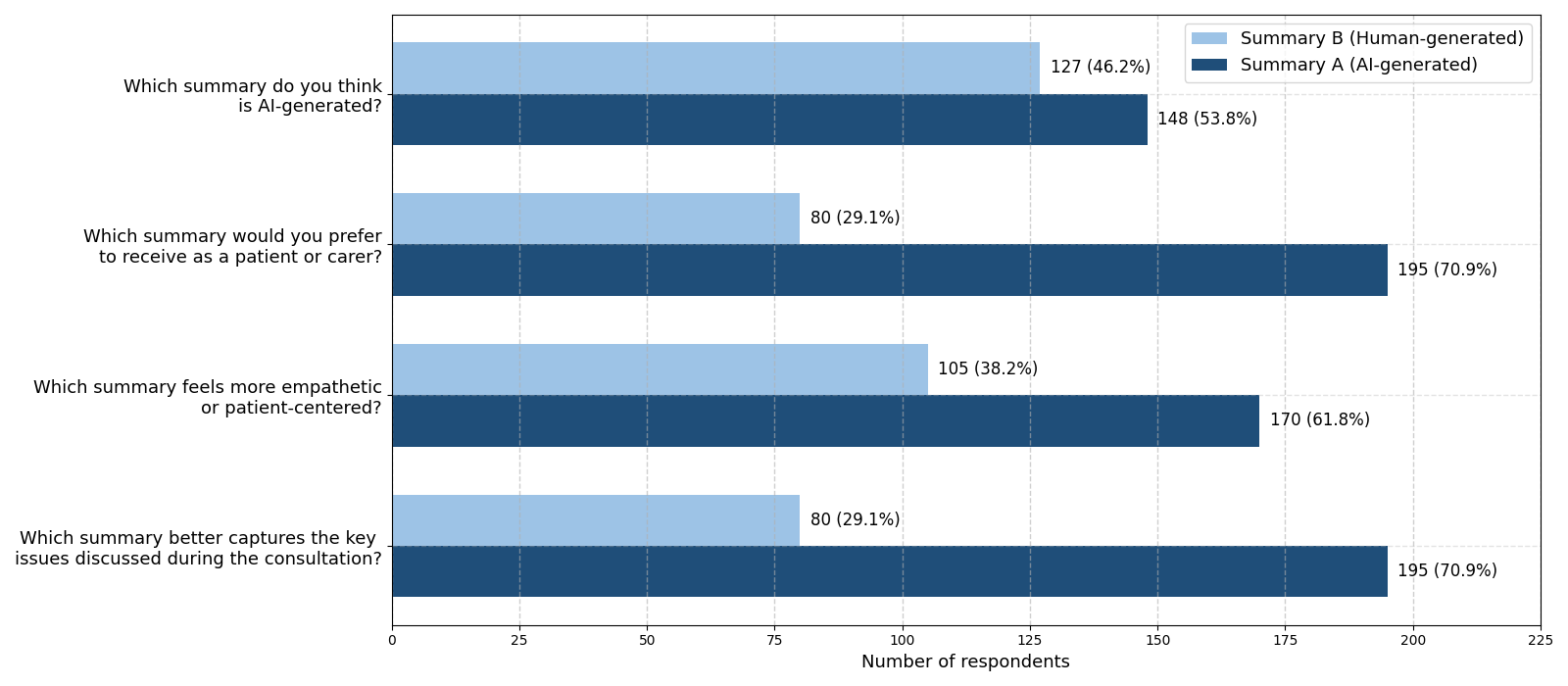}
    \caption{Voting results for Summary A (AI) vs. Summary B (human)}
    \label{fig_summary}
\end{figure*}

The evaluation results are shown in Figure~\ref{fig_summary}. For three of the four questions, Summary A (AI) was strongly favoured.
\begin{itemize}
    \item A total of \textbf{195} participants (\textbf{70.9\%}) judged Summary A (AI) as better capturing the key issues discussed during the consultation, compared to \textbf{80} participants (\textbf{29.1\%}) who selected Summary B (human).
    \item Similarly, \textbf{170} participants (\textbf{61.8\%}) rated Summary A (AI) as more empathetic or patient-centred, while \textbf{105} participants (\textbf{38.2\%}) selected Summary B (human).
    \item When asked which summary they would prefer to receive as a consumer, \textbf{195} participants (\textbf{70.9\%}) chose Summary A (AI) and \textbf{80} participants (\textbf{29.1\%}) chose Summary B (human).
\end{itemize}

The qualitative responses explain why Summary A (AI) was consistently preferred across these dimensions, while also clarifying why a substantial minority still favoured Summary B (human). 

\begin{itemize}
    \item When judging which summary better captured the key issues and which they would prefer to receive, participants applied different and sometimes competing criteria. Some valued relevance and conciseness, noting that Summary A (AI) ``\textit{focuses on the key issues in the consultation better by omitting some unnecessary details}'' and was ``\textit{straight to the point}'' or ``\textit{easier to get the key information from}''. Others preferred greater completeness, observing that there were ``\textit{more details in summary B compared to A}''. This indicates that the quantitative preference split reflects trade-offs between relevance, completeness, and cognitive usability rather than disagreement about the task itself.
    \item The strong quantitative difference in perceived empathy is also explained by participants’ qualitative comments. \textbf{Judgements of empathy were driven primarily by linguistic and stylistic cues rather than clinical content}. Participants described Summary A (AI) as more empathetic when the ``\textit{tone is a bit warmer, less clinical}'', when it ``\textit{actually uses people’s names}'', or when it avoids harsh or impersonal phrasing. Even subtle lexical choices influenced perception, as illustrated by the comment that ``\textit{B uses `suffers from', whereas A uses `has'}''. These observations help explain why empathy ratings diverged more strongly than some of the other measures: participants were highly sensitive to narrative voice, tone, and framing.
\end{itemize}

In contrast, the responses were split more evenly when participants were asked to identify which summary they believed was AI-generated. A total of \textbf{148} participants (\textbf{53.8\%}) selected Summary A (AI), while \textbf{127} participants (\textbf{46.2\%}) selected Summary B (human). Qualitative responses suggest that participants relied primarily on stylistic cues rather than factual correctness, referring to features such as ``\textit{it feels like it follows a template}'', ``\textit{feels more emotionless}'', or ``\textit{sounds like it has been written by AI}''. These judgements were rarely grounded in perceived clinical errors, but instead in tone, uniformity of structure, and emotional flatness. This explains why, despite strong preferences and perceived differences in quality and empathy, participants were less certain and less consistent in identifying which summary was AI-generated.

Overall, preference for Summary A (AI) remained strong regardless of perceived authorship. Among participants who correctly believed Summary A (AI) was AI-generated, the majority still rated it as better capturing the key issues and more empathetic. Similarly, even among those who believed Summary B (human) was AI-generated, Summary A (AI) was still more frequently preferred. This suggests that participants’ evaluations were driven more by perceived quality and tone than by assumptions about whether the summary was written by AI or a clinician.

\subsection{Qualitative reflections on trust, safety, and the broader impact of AI in healthcare}

In addition to the scenario-based comparison task, participants answered three open-ended questions concerning (i) \textit{what would make AI tools safer or more trustworthy in healthcare}, (ii) \textit{how they believed AI might impact their healthcare experience}, and (iii) \textit{any further thoughts, concerns, or suggestions}. We conducted an exploratory qualitative analysis of these free-text responses using an unsupervised, data-driven clustering approach. 

Although participants’ responses exhibited substantial lexical and semantic diversity, the clusters consistently organised around a small number of higher-order dimensions: (i) trust, safety, and governance, (ii) human oversight and accountability, (iii) efficiency and system-level impact, and (iv) the preservation of human connection and care. These dimensions provide important explanatory context for the quantitative findings, which showed neither uniformly positive nor uniformly negative attitudes towards AI in healthcare.

\subsubsection{What would make AI tools safer or more trustworthy}

The quantitative results showed strong endorsement of items emphasising safety, oversight, and responsible use of AI in healthcare. The open-ended responses clarify that participants conceptualise safety and trustworthiness primarily as socio-technical and regulatory issues rather than purely technical ones \cite{zhou2025framework}.

Participants repeatedly called for governance mechanisms such as clinician review, professional accountability, and healthcare system's responsibility. Typical comments included: ``\textit{If the doctor reviews them [AI generated summary] before they are available to the patient}'', ``\textit{Healthcare provider should review all the AI-generated summary}'', and ``\textit{Having the information validated by a human}''. Others emphasised transparency, auditability, and clear responsibility in case of errors. These responses closely mirror the quantitative preference for human-in-the-loop arrangements and indicate that trust in AI is conditional on visible and credible oversight structures.

\subsubsection{How participants expect AI to impact their healthcare experience}

When asked how AI might affect their healthcare experience, participants expressed a mixture of optimism and concern, consistent with the mixed pattern observed in the quantitative results.

On the positive side, many participants anticipated improvements in efficiency, documentation quality, and clinician workload, with potential flow-on benefits for access and continuity of care. 

On the negative side, others worried about depersonalisation and loss of human connection, for example, noting that ``\textit{AI does not have the human aspect -- \textbf{it is hard to effectively convey your issues to a computer}}'' or that they ``\textit{just feel more comfortable with a human that might have personal experience with what I'm going through}''. These responses show that participants are weighing potential system-level gains against possible impacts on relational and experiential aspects of care.

\subsubsection{Additional concerns and suggestions}

In the final open-ended question, participants raised a range of further issues that span across trust, governance, and experience. These included concerns about over-reliance on AI, erosion of clinical responsibility, data privacy, and the risk that \textit{AI involvement could be used to reduce human contact rather than support it}. Others reiterated the importance of positioning \textit{AI as a supportive tool rather than a replacement for clinicians}, and of ensuring that \textit{patients remain informed and able to question or challenge AI-mediated outputs}.

Taken together, these qualitative reflections provide a coherent explanatory layer for the quantitative findings. While the survey ratings reveal overall patterns of cautious, conditional acceptance of AI in healthcare, the free-text responses show that participants’ attitudes are structured around a small number of interrelated concerns: safety and governance, human oversight, efficiency, and the preservation of human connection in care. This confirms that consumer attitudes towards AI in healthcare are not reducible to a single dimension of trust or acceptance, but instead reflect a multi-dimensional evaluative framework shaped by both practical considerations, and social and ethical expectations.

\section{Discussion}\label{discussion}

This study provides a mixed-method examination of how consumers evaluate the use of AI in healthcare, with a particular focus on AI-generated consultation summaries. By combining general attitudinal measures, a scenario-based comparison of realistic-simulation patient-facing artefacts, and qualitative explanations of participants’ reasoning, the study extends prior work that has primarily examined public attitudes towards clinical AI at an abstract level \cite{young2021patient, nuccetelli2025use, foresman2025patient}. Our findings show that consumer attitudes towards healthcare AI are neither uniformly positive nor uniformly negative, but instead structured around multiple, partially independent dimensions including trust, communication quality, empathy, and governance. This multi-dimensional structure helps explain why acceptance of AI in healthcare has proven uneven and contested despite rapid technical progress \cite{kwong2024integrating, silcox2024potential}.

\subsection*{Trust in healthcare AI is conditional}

Throughout the survey, trust in AI clustered around moderate levels. For example, 32.0\% of the participants reported low trust and only 16.4\% reported high trust for general trust level of AI in health. This pattern closely mirrors findings from large-scale reviews and qualitative studies showing that consumers tend to adopt a cautious, conditional stance towards clinical AI rather than either enthusiastic acceptance or outright rejection \cite{young2021patient, esmaeilzadeh2021patients, gundlack2025patients}. In our data, this ambivalence was not a sign of confusion or lack of opinion, but reflected a nuanced evaluation of risks, benefits, and responsibilities.

The qualitative findings clarify that participants did not treat trust as a property of the technology alone. Instead, trust was framed as a judgement mediated by healthcare systems, professional responsibility, and visible human oversight mechanisms. This is consistent with psychological and behavioural evidence that trust in AI systems depends heavily on perceived epistemic authority, accountability, and social context rather than on technical performance alone \cite{ding2025trust, reis2025public}. Participants’ repeated emphasis on clinician review and healthcare systems' responsibility suggests that, in healthcare, trust is anchored less in algorithms than in the social systems that govern their use.

Taken together, these findings reinforce the view that trust in healthcare AI is best understood as conditional, relational, and context-dependent. This has important implications for technology development, deployment strategies, and regulatory evaluation, suggesting that efforts to increase trust should focus not only on improving models, but also on strengthening governance, communication, and professional accountability structures.

Although this study was conducted in Australia, the findings are particularly relevant given the country’s rapid expansion of telehealth services, geographically dispersed population, and active policy focus on trustworthy AI in healthcare. Australia has introduced national frameworks emphasising safety, accountability, and human oversight in clinical AI deployment \cite{aus_ai_ethics, aus_ai_assurance}. At the same time, the conditional trust patterns observed here are consistent with international evidence, suggesting broader applicability beyond the Australian context.

\subsection*{Consumers evaluate concrete AI outputs}

A central contribution of this study is that it examines how consumers respond to a concrete, patient-facing AI artefact rather than to AI in the abstract. When participants evaluated two realistic-simulation consultation summaries, they showed strong and consistent preferences across multiple dimensions, yet were far less consistent in identifying which summary was AI-generated. This dissociation suggests that, in practice, consumers judge AI systems primarily through the quality of the outputs they encounter, rather than through prior beliefs about AI involvement \cite{reis2024influence, changes_chatgpt_perception}.

This finding complements recent clinical and translational work on AI-based summarisation, which has largely focused on technical accuracy, completeness, and efficiency \cite{chen2024exploring, clough2024transforming, shemtob2025comparing}. Our results indicate that, from a consumer perspective, these technical metrics are only part of the evaluative picture. Relevance, clarity, and perceived empathy cues play a central role in shaping acceptance and preference, even when clinical content is generally similar.

More broadly, this suggests that patient-facing AI systems should be evaluated not only as clinical tools, but also as communication technologies. This perspective aligns with growing recognition that the success of AI in healthcare depends as much on how outputs are experienced and interpreted as on how they are generated \cite{kwong2024integrating, silcox2024potential}.

\subsection*{Empathy is driven by tone and framing}

One of the most striking findings in this study is the magnitude of the difference in perceived empathy between the two summaries. The qualitative analysis shows that these judgements were driven primarily by tone, narrative voice, and lexical choices rather than by differences in clinical content. This is consistent with recent experimental work demonstrating that perceived empathy in medical communication - whether produced by clinicians or AI systems - is highly sensitive to stylistic and framing cues \cite{chen2025patient, ovsyannikova2025third}.

This sensitivity has important implications. It suggests that even technically accurate and complete AI-generated summaries may be experienced as cold, impersonal, or alienating if their linguistic form does not meet consumers’ expectations for care and respect. Conversely, it also raises the possibility that AI systems may be able to meet or even exceed baseline expectations for empathic communication under some conditions, a prospect that carries both opportunities and ethical challenges \cite{goodman2024ai}.

At a more general level, these findings reinforce arguments that evaluation of clinical AI text should extend beyond correctness and safety to include communicative and relational quality. In patient-facing contexts, the way information is said matters as much as what is said.
Conversely, care should be taken that empathy cues are not misused in AI technology design to try and hide deficiencies in accuracy and privacy.

\subsection*{Gap between governance and lived experience}

Across both quantitative and qualitative components of the study, the most consistent and strongest signal concerned governance rather than capability. Participants expressed clear expectations that clinicians should remain responsible for AI outputs ($\mathbf{90.9\%}$ agreed), that review should occur before summaries are shared ($\mathbf{88\%}$ agreed), and that accountability structures should be explicit ($\mathbf{86.9\%}$ agreed). These expectations align closely with international and Australian policy frameworks that emphasise human oversight, accountability, and transparency \cite{who_ai_health, oecd_ai, aus_ai_ethics, aus_ai_assurance, tga_ai_md}.

However, the qualitative findings also reveal an important gap. While regulatory frameworks tend to frame safety primarily in terms of technical validation, robustness, and compliance \cite{tga_ai_md, aus_ai_assurance}, participants framed safety and trust in experiential and relational terms: whether outputs are understandable, respectful, and clearly owned by a responsible professional. This echoes broader arguments that trustworthy AI in healthcare is as much an organisational and social challenge as a technical one \cite{lekadir2025future, kwong2024integrating, saez2024resilient}.

This gap suggests that current regulatory approaches, while necessary, may not be sufficient to secure social legitimacy for patient-facing AI systems. Incorporating consumer experience and perception into assurance and evaluation frameworks may therefore be critical for sustainable and acceptable deployment.

\subsection*{Implications for regulation and practice}

Several implications follow from these findings. First, evaluation and assurance frameworks should expand beyond technical accuracy and clinical risk to include patient-centred criteria such as clarity, tone, and perceived respectfulness \cite{goodman2024ai, kwong2024integrating}. Second, governance arrangements should ensure not only that human oversight exists, but that it is visible and meaningful to consumers, reinforcing trust and accountability \cite{lekadir2025future, saez2024resilient}. Third, policies should ensure that consumers retain transparency and meaningful choice about AI’s role in their care, consistent with international and Australian guidance \cite{who_ai_health, aus_ai_ethics, aus_ai_assurance}. Finally, consumer participation should be embedded more systematically into the design, evaluation, and regulation of healthcare AI, echoing calls from participatory and implementation-focused research \cite{foresman2025patient, silcox2024potential}.
This is especially relevant for Australia’s telehealth context, where AI summaries may support care continuity across remote and urban settings.

\subsection*{Limitations and directions for future research}

The scenario part of the survey relies on a single clinical scenario and two summaries, which limits generalisability across specialties, contexts, and writing styles. 
However, including multiple scenarios was not practical, as it would lengthen the survey risking recruitment and attrition.
The online sample may also under-represent populations with lower digital access or health literacy, a limitation shared by many studies in this area \cite{young2021patient, nuccetelli2025use}. Participants evaluated hypothetical scenarios rather than experiencing AI systems in real clinical workflows, which may affect real-world applicability.
In addition, as with many online surveys conducted in Australia, the sample was somewhat more highly educated and metropolitan than the national census distribution.

An additional limitation concerns the relational and communicative gap between human care and AI systems. Several participants noted that “\textit{AI does not have the human aspect}” and that \textit{it may be difficult to convey personal concerns effectively to a computer}. This highlights that the effectiveness of AI summarisation depends not only on model performance, but also on the quality of patient–clinician interaction, contextual understanding, and the preservation of human connection in care \cite{kwong2024integrating,foresman2025patient}.

Future work should examine a broader range of clinical contexts, conduct longitudinal study in real clinical settings, and investigate how repeated exposure to AI-generated documentation reshapes trust, expectations, and patient - clinician relationships over time \cite{kwong2024integrating, silcox2024potential}. Further research is also needed on cultural, linguistic, and accessibility dimensions of patient-facing AI communication, examining how AI tools can support, rather than erode, the interpersonal dimensions of healthcare communication.

\section{Methods}\label{method}

\subsection{Study design}

We conducted a cross-sectional, mixed-methods online survey, deployed on Qualtrics\footnote{\url{https://www.qualtrics.com}}, to examine consumer attitudes towards the use of AI in healthcare in Australia, with a specific focus on trust, perceived quality, empathy, and preferences regarding AI-assisted clinical consultation summaries. The study combined structured quantitative items with open-ended qualitative questions in order to capture both population-level attitudinal patterns and the reasoning underlying participants’ judgments.

The survey also included a scenario-based comparison task in which participants evaluated two alternative versions of a clinical consultation summary (authored by AI and  a human clinician, undisclosed to the participants which is which), followed by closed and open questions.

\subsection{Ethics}

The study protocol, survey instrument, and participant information materials were reviewed and approved by the University Human Research Ethics Committee (HREC)\footnote{This project was approved by Monash University HREC (\#49238).}. 

The doctor-patient conversation used to generate two summaries was carefully designed by a clinical expert to mimic authentic interactions that occur between patients and healthcare professionals.
Therefore, no personally identifiable information (PII) or sensitive patient data was included in the survey content, which adheres to ethical guidelines and legal regulations in health care research.

The survey was anonymous, and no identifying personal data were collected.

\subsection{Participants and recruitment}

After gaining ethics approval, participants were recruited via the Prolific\footnote{\url{https://www.prolific.com}} online research platform and social media channels (e.g., LinkedIn). 
Prolific’s built-in eligibility filters and attention checks were used to support response quality and authenticity.

Eligibility criteria included being aged 18 years or older, residing in Australia, and being able to read and understand English. 
These criteria were selected to ensure that participants reflected the intended population of healthcare consumers in routine care, and that all participants could meaningfully evaluate the scenario materials within the Australian healthcare context. 
No medical background was required, as the study aimed to capture consumer perspectives rather than clinician evaluations. All participants provided informed consent before beginning the survey.

\subsection{Survey instrument and procedure}

The survey instrument consisted of multiple sections combining demographic questions, standardised attitudinal scales, a scenario-based comparison task, and open-ended reflection questions.

First, participants completed demographic and background questions, including age, gender, location, education level, self-rated digital technology skill,  general attitude towards new technology, and general trust level towards AI in healthcare.

Second, participants completed a set of Likert-scale items measuring technology readiness towards AI in healthcare, adapted from the Technology Readiness Index (TRI) \cite{khoza2024technology}, covering optimism, innovativeness, discomfort, and insecurity dimensions.

Third, participants were presented with a short, realistic-simulation consultation scenario followed by two summaries, one generated by AI and one written by a clinician, although it was not disclosed to participants which is which. Participants were asked four questions about quality, empathy, preference, and identification of the AI-generated summary. For each question, participants were also asked to briefly explain their reasoning in free text.

Fourth, participants completed a set of Technology Acceptance Model (TAM) \cite{fam2025modeling} items measuring perceived usefulness and perceived ease of use of AI generated summaries in healthcare, followed by items measuring overall attitudes and behavioural intention towards using AI-generated summaries.

Next, participants completed a set of Likert-scale items assessing risk, trust, and ethical concerns related to AI in healthcare, including concerns about data use, accuracy, responsibility, transparency, and trust impacts.

Finally, participants answered open-ended questions about what would make AI tools safer or more trustworthy in healthcare, how they believed AI might impact their healthcare experience, and whether they had any additional thoughts or concerns.

The order of presentation was fixed to ensure all participants evaluated the same materials in the same context. The full survey took approximately 10 - 15 minutes to complete.

\subsection{Quantitative measures and analysis}

All closed-ended items, with the exception of the scenario-based comparison questions, were measured using 5-point Likert scales. For analysis and visualisation, Likert responses were linearly transformed to a numeric scale ranging from $-2$ (Strongly disagree) to $+2$ (Strongly agree), with $0$ representing a neutral response.

The quantitative instrument comprised five domains: (i) general technology readiness towards AI in healthcare, (ii) perceived usefulness and ease of use of AI for summarization, (iii) overall attitudes and behavioural intention towards AI-generated summaries, (iv) perceived risks, trust, and governance expectations, and (v) scenario-based comparative judgments between the two summaries. The scenario-based questions required participants to make a forced-choice selection between the two summaries for each of four evaluative dimensions and were therefore treated as categorical choice outcomes rather than as scaled ratings.

In line with the exploratory aims of the study, all items were analysed and reported at the item level rather than aggregated into composite scores, in order to preserve interpretability and to facilitate integration with the qualitative findings.

Descriptive statistics (means, percentages, and response distributions) were computed for all quantitative items. Given the descriptive and exploratory nature of the study, analyses focused on characterising population-level patterns rather than formal hypothesis testing. All quantitative analyses were conducted using Python.

\subsection{Qualitative data and analysis}

The survey included eight open-ended response fields per participant, resulting in a corpus of free-text responses explaining participants’ judgments and perspectives.
These questions captured (i) participants' reasoning for the general trust level of AI in healthcare, (ii) participants’ reasoning for their comparative evaluations of AI generated summary versus clinician-written summary (quality, empathy, preference, and perceived authorship), and (iii) broader reflections on what would make AI tools safer or more trustworthy in healthcare, how AI might shape future care experiences, and any additional concerns or suggestions. 

We conducted an exploratory, data-driven qualitative analysis using unsupervised text clustering. For each open-ended question, responses were embedded using sentence-level semantic embeddings and clustered using K-means clustering. A fine-grained solution with $K = 9$ clusters per question was used to surface diverse sub-themes. These clusters were then examined and interpreted by the research team and synthesised into higher-level conceptual themes.

\subsection{Integration of quantitative and qualitative analyses}

Quantitative and qualitative findings were integrated at the interpretation stage using a convergent mixed-methods approach \cite{storey2025guiding}. Quantitative results were first used to establish overall patterns of trust, preference, empathy, and concern. Qualitative findings were then used to explain these patterns, clarify sources of disagreement or variance, and reveal the criteria participants used in forming their judgments.

% \section{Conclusion}\label{conclusion}

% This study provides a consumer-centred examination of how people evaluate the use of AI in healthcare when it is encountered not as an abstract concept, but as a concrete, patient-facing artefact in the form of a consultation summary. By combining general attitudinal measures, a scenario-based comparison of realistic-simulation summaries, and qualitative analysis of participants’ reasoning, the study shows that consumer attitudes towards healthcare AI are structured around multiple, partially independent dimensions rather than a single axis of acceptance or rejection.

% The results also highlight an important gap between current regulatory and assurance frameworks, which focus primarily on technical validation and compliance, and consumer expectations, which are framed in experiential, relational, and regulatory terms. For patient-facing generative AI systems, safety and trust are not only matters of correctness, but also of communicative quality, respectfulness, and visible human oversight.

% Taken together, these findings suggest that the sustainable and socially legitimate deployment of AI in healthcare will require evaluation and governance approaches that integrate patient experience alongside technical performance. More systematic inclusion of consumer perspectives in design, implementation, and regulatory assessment will be essential if AI systems are to support, rather than undermine, trust, understanding, and the human dimensions of care.

\section*{Acknowledgments}

This study is supported by the Digital Health Cooperative Research Centre Limited (DHCRC) and Monash University. DHCRC is funded under the Commonwealth Government Cooperative Research Centres Program.

\section*{Declaration of conflicting interests}
% Joycelyn Ling is the Senior Program Manager of Digital Health CRC.
The authors declared no potential conflicts of interest with respect to the research, authorship, and/or publication of this article.

\section*{Funding}
This work was supported by Digital Health CRC Limited (DHCRC) under grant DHCRC-0334. 

\section*{Ethical approval and informed consent statements}
This project was approved by Monash University Human Research Ethics Committee (\#49238). 
Every participant gave informed consent prior to participation.

\section*{Data availability}
The datasets collected and/or analyzed during the current study are not publicly available, but the data are available from the corresponding author on reasonable request.

\section*{Code availability}
The underlying code for this study is not publicly available but may be made available to qualified researchers on reasonable request from the corresponding author.

\section*{Author contributions}
WZ, RH, and JL conceived the study. 
RH secured funding and supervised the research. 
WZ wrote the draft.
All authors critically revised the manuscript for important intellectual content. 
All authors provided comments and approved the paper.

\bibliographystyle{nature}
\bibliography{ref}


\end{document}